\documentclass[lettersize,journal]{IEEEtran}
\usepackage{amsmath,amsfonts}
\usepackage{algorithmic}
\usepackage{booktabs}
\usepackage{multirow}
\usepackage{algorithm}
\usepackage{array}
\usepackage[caption=false,font=normalsize,labelfont=sf,textfont=sf]{subfig}
\usepackage{textcomp}
\usepackage{stfloats}
\usepackage{url}
\usepackage{verbatim}
\usepackage{graphicx}
\usepackage[table]{xcolor} 
\usepackage{cite}
\usepackage[pscoord]{eso-pic}
\newcommand{\placetextbox}[3]{
\setbox0=\hbox{#3}
\AddToShipoutPictureFG*{ \put(\LenToUnit{#1\paperwidth},\LenToUnit{#2\paperheight}){\vtop{{\null}\makebox[0pt][c]{#3}}}
}
}
\placetextbox{.5}{0.055}{\small{DISTRIBUTION STATEMENT A. Approved for public release; distribution is unlimited. OPSEC\# 9830.}}

\begin{document}

\title{Anomalous Decision Discovery using Inverse Reinforcement Learning}

\definecolor{lightgray}{gray}{0.90}  

\author{
  Ashish~Bastola\IEEEauthorrefmark{1},~\IEEEmembership{Student Member,~IEEE,}
  Mert~D.~Pesé\IEEEauthorrefmark{1},~\IEEEmembership{Member,~IEEE,}
  Long~Cheng\IEEEauthorrefmark{1},~\IEEEmembership{Member,~IEEE,}
  Jonathon~Smereka\IEEEauthorrefmark{2},
  and~Abolfazl~Razi\IEEEauthorrefmark{1},~\IEEEmembership{Senior Member,~IEEE}
  \thanks{Manuscript received Month Day, Year; revised Month Day, Year.}
  \thanks{\IEEEauthorrefmark{1}Ashish Bastola, Mert D. Pesé, Long Cheng, and Abolfazl Razi are with the School of Computing, Clemson University, Clemson, SC, USA 
  (e-mail: \{abastol, mpese, lcheng2, arazi\}@clemson.edu).}
  \thanks{\IEEEauthorrefmark{2}Jonathon Smereka is with U.S. Army DEVCOM GVSC (jonathon.m.smereka.civ@army.mil)}
}

\maketitle

\begin{abstract}
Anomaly detection plays a critical role in Autonomous Vehicles (AVs) by identifying unusual behaviors through perception systems that could compromise safety and lead to hazardous situations. Current approaches, which often rely on predefined thresholds or supervised learning paradigms, exhibit reduced efficacy when confronted with unseen scenarios, sensor noise, and occlusions, leading to potential safety-critical failures. Moreover, supervised methods require large annotated datasets, limiting their real-world feasibility. To address these gaps, we propose an anomaly detection framework based on Inverse Reinforcement Learning (IRL) to infer latent driving intentions from sequential perception data, thus enabling robust identification. Specifically, we present Trajectory-Reward Guided Adaptive Pretraining (TRAP), a novel IRL framework for anomaly detection, to address two critical limitations of existing methods: noise robustness and generalization to unseen scenarios. 
Our core innovation is implicitly learning temporal credit assignments via reward and worst-case supervision. We leverage pretraining with variable-horizon sampling to maximize time-to-consequence, resulting in early detection of behavior deviation. Experiments on 14,000+ simulated trajectories demonstrate state-of-the-art performance, achieving 0.90 AUC and 82.2\% F1-score - outperforming similarly trained supervised and unsupervised baselines by 39\% on Recall and 12\% on F1-score, respectively. Similar performance is achieved while exhibiting robustness to various noise types and generalization to unseen anomaly types. Our code will be available at: https://github.com/abastola0/TRAP.git
\end{abstract}


\section{Introduction}
Autonomous Vehicles (AVs) represent a significant advancement in transportation systems. However, their safe deployment greatly depends on robust anomaly detection mechanisms to address unpredictable real-world scenarios \cite{solaas2024systematic, nazat2024evaluating}. While AI-driven perception systems excel in controlled environments, their reliability diminishes when encountering novel edge cases—a critical gap given that 67\% of AV incidents stem from undetected behavioral anomalies in safety-critical situations \cite{solaas2024systematic, bogdoll2022anomaly}. Traditional anomaly detection methods face three fundamental limitations: (i) Dependence on manually crafted rules that fail to generalize across driving contexts in case of rule-based methods \cite{li2022outlier, nazat2024evaluating}, (ii) Model-based supervised approaches that generally require exhaustive labeled datasets are impractical for rare events \cite{rajendar2022sensor, vajda2024machine}, and (iii) inability of unsupervised methods to distinguish safety-critical deviations from benign outliers in sequential decision-making processes \cite{datacamp_anomaly_detection}.

Recent advances in Deep Learning (DL) methods have enabled data-driven anomaly detection through architectures such as Long Short-Term Memory (LSTM), LSTM-based autoencoders \cite{wang2024anomaly}, and multi-modal sensor fusion \cite{bogdoll2022anomaly} techniques. However, these methods face scalability issues with high-dimensional data, often requiring additional pre-processing steps like Principal Component Analysis (PCA) to reduce dimensionality, which adds to the risk of losing nuanced features. Moreover, threshold tuning remains dataset-specific and labor-intensive, which is worsened by their interpretability issues as they mostly perform full trajectory-wise classification (assign labels to entire trajectory) \cite{nazat2024evaluating}.

Inverse Reinforcement Learning (IRL) addresses this by inferring reward functions from expert demonstrations, providing a principled framework for understanding decision-making patterns, facilitating anomaly detection \cite{oh2019sequential, li2022outlier}. While prior IRL-based approaches \cite{oh2019sequential} demonstrated promise in controlled simulations, they often suffer from three key shortcomings: 

\begin{itemize}
    \item \textbf{Threshold Dependency}: Manual tuning of normality thresholds limits adaptability to diverse driving scenarios, 
    \item \textbf{Temporal Myopia}: Temporal Myopia refers to bias when the model prioritizes more on immediate reward and short-term gains over long-term consequences. Fixed-horizon optimization fails to account for the delayed consequences of anomalous actions, 
    \item \textbf{Sensor Noise Vulnerability}: Performance degradation under real-world noise conditions like rain and sensor dropout.
\end{itemize}

Our framework introduces three contributions to overcome these limitations:

\begin{itemize}
    \item  We introduce TRAP (Trajectory-Reward Guided Adaptive Pretraining), a novel IRL-based anomaly detection framework that implicitly learns temporal credit assignment via reward and worst-case supervision, eliminating the need for extensively labeled datasets. Given expert demonstrations and worst-case rollouts, our method leverages variable horizon (unequal episode length) pretraining to detect behavioral deviations by implicitly maximizing the time to consequence. 
    \item We formulate rule-based anomaly labeling criteria to clearly define the joint anomaly threshold. This labeling allows us to evaluate our model's performance in identifiable real-world anomalous scenarios such as tailgating, sudden braking, etc. Note that this labeling is only for validation, and we do not use this labeled supervision during training. We assume our method is generalizable to broad anomaly groups, with our labeled groups being a small subset that can be used for deterministic validation. 
    \item We generate a closed-loop simulation eligible anomaly trajectory dataset that can be used for various RL-based and supervised applications. Since we perform anomaly labeling purely based on the vehicle's behavior in that specific trajectory after simulation, we prevent any kind of bias in the dataset generation. The contributed anomaly dataset—14,000+ trajectories with six anomaly categories—provides a benchmark for evaluating safety-critical AV behaviors under 8 noise conditions, addressing the scarcity of realistic scenarios noted before. 
\end{itemize}

We validate our approach through large-scale simulations in the MetaDrive \cite{li2022metadrive} simulator due to its lightweight and high-speed performance. Our method demonstrates 12.5\% higher recall than unsupervised baselines and 39\% over supervised methods that are benchmarked on our dataset. 

\section{Related Work}
 Anomaly detection originates from the broader challenge of outlier detection. In many cases, the datasets used for outlier detection are static, i.e., the data does not represent a sequence that evolves over time, but rather a fixed collection of observations.
 Various methods exist for detecting outliers, including supervised techniques \cite{abe2006outlier, chu2020neural}, distance-based approaches \cite{angiulli2009dolphin, knorr2000distance}, feature selection-based methods \cite{zhang2024realnet}, density-based methods \cite{breunig2000lof, hu2018anomaly, deng2023bootstrap}, model-based strategies \cite{he2003discovering}, and isolation-based techniques \cite{liu2008isolation}. However, few methods have evaluated their feasibility over vehicle trajectory data to detect \cite{oh2019sequential} sequentially occurring anomalous behaviors.

\subsection{Inverse Reinforcement Learning for Anomaly Detection}
    
IRL aims to identify the underlying intent or the goal the agent is trying to reach by calculating intermediate reward estimates for each agent's transition. The reward estimates are commonly modeled as a distribution using neural networks, which are gradually optimized using expert demonstrations. Applying IRL for anomaly detection is one of the theoretically appropriate ways to detect anomalies \cite{oh2019sequential}. The idea is simple: first, estimate the reward baseline and specify a threshold for some deviation measure of a test trajectory from the reward distribution. One such method leveraging this approach for sequential data using Guided Cost Learning (GCL) \cite{finn2016guided} for reward learning includes a sequential anomaly detection framework \cite{oh2019sequential} which uses policy optimization in online Reinforcement Learning (RL) settings. Using online environment interaction as part of online RL is significantly time-consuming, sample inefficient, unstable, and poses major computational limitations when dealing with training and simulation of high-dimensional data simultaneously. 
Moreover, pure online learning methods operate in a strictly forward manner without access to past trajectories, which limits the ability to apply strategies that optimize for longer time to anomaly detection.
    An IRL-based approach has also been used in diverse scenarios, such as Unmanned Aerial Vehicles (UAV), by providing a correction mechanism for the flight anomalies \cite{lian2022anomaly}. Li \textit{et al.} \cite{li2022outlier} developed an outlier-robust IRL for detecting anomalous driving behaviors. Their joint IRL framework enables learning a common reward function that captures the behavior of the majority while simultaneously identifying outliers. All of these methods are, however, threshold-based and require threshold tuning to fit each specific application. Most of the code is not publicly released, and thus, cross-comparison is challenging; however, we reproduced the work by Oh \textit{et al.} \cite{oh2019sequential}, which is the most popular one for the purpose of comparison with our method. 

    \subsection{Benchmarking and Dataset Considerations}
    Developing suitable benchmarks is essential for assessing anomaly detection methods. However, most existing sequential anomaly benchmarks focus on objects, often overlooking anomalies in vehicle behavior. Bogdoll \textit{et al.} \cite{bogdoll2024anovox} introduced AnoVox, described as the largest benchmark for anomaly detection in autonomous driving up to date. Their benchmark incorporates large-scale multimodal sensor data and spatial voxel ground truth, allowing for comparing methods independent of the sensors used. However, this benchmark \cite{bogdoll2024anovox} focuses more on anomalous objects than anomalous behavior like sudden turns, lane crossing, etc. The next limitation is the use of the CARLA simulator, which operates at a significantly lower frame rate than the MetaDrive simulator \cite{li2022metadrive}, thereby posing a substantial constraint on both training and evaluation when comparing different baselines. Moreover, CARLA faces severe compatibility issues with recent Linux distributions \cite{carla_discussion_6237}, thus making it less feasible. In our case, since we need to generate Worst-Case Terminus (WCT), which is the terminating condition leading to hazards, we need a large number of simulations for both expert and test trajectories and thus opt for the significantly more lightweight MetaDrive simulator, which offers satisfactory simulation quality without sacrificing speed. Other similar approaches like Fontanel \textit{et al.} \cite{fontanel2021detecting} utilize semantic segmentation to detect anomalous events such as traffic light abnormalities, unusual obstacles, hazards, etc. However, these can be formulated as general image recognition problems and are different from behavioral anomaly detection compared to ours.

    \subsection{Rule-based and symbolic approaches}
    Beyond purely data-driven methods, rule-based approaches offer complementary strengths in anomaly detection. For instance, Chen \textit{et al.} \cite{chen2019augmenting} explored integrating a common-sense knowledge base and predefined rules on driving behavior to enhance anomaly detection in autonomous vehicles. By leveraging symbolic rules, this approach mitigates the limitations of incomplete training data while enabling dynamic rule updates in an active learning fashion.
    In automotive cybersecurity, numerous attacks have exploited vulnerabilities in Controller Area Networks (CAN), the communication backbone between Electronic Control Units (ECUs) \cite{nie2018over, nie2017free}. To counter such threats, rule-based methods such as One-Class Anomaly Detection Systems (OCADS) and association rules have been proposed to detect unknown and adversarial attacks \cite{kumar2025orads, d2023association}. However, we focus on detecting anomalies as high-level consequences of these attacks, allowing us to capture behavioral deviations indicative of a broader range of security breaches.
    
    While rule-based thresholding can be useful in directly observable feature spaces, its broader application remains limited. Our TRAP framework primarily relies on Inverse Reinforcement Learning (IRL). Yet, it integrates rule-based components to define worst-case termination scenarios, aligning with the conceptual foundation of rule-driven anomaly detection. This hybrid approach can extend to various anomaly types. However, purely rule-based methods often struggle with unseen cases, making them more suited as heuristic guides for trend identification or as complementary safeguards alongside more robust detection mechanisms.

    \begin{figure}[t]
        \centering
        \includegraphics[width=0.9\linewidth]{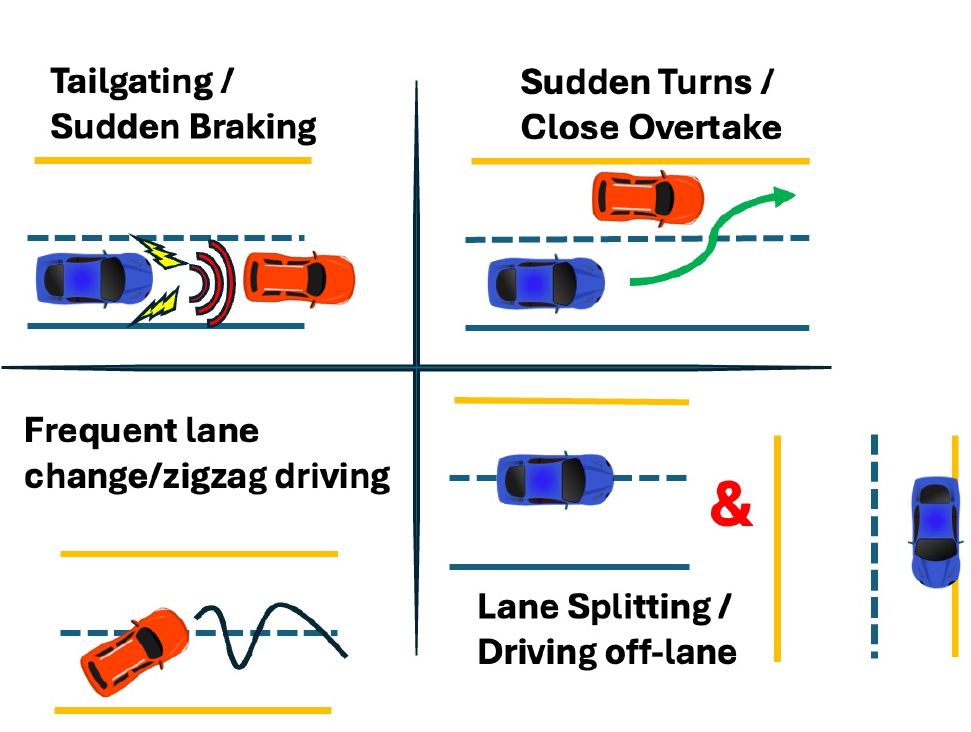}
        \caption{Anomalies classified using a rule-based approach. We classify common anomaly scenarios for evaluation, such as sudden braking, frequent lane change, etc., by analyzing vehicle kinematics and environment information from the simulator. }
        \label{fig:anomaly_types}
    \end{figure}
    
    \begin{figure*}
        \centering
        \includegraphics[width=0.9\linewidth]{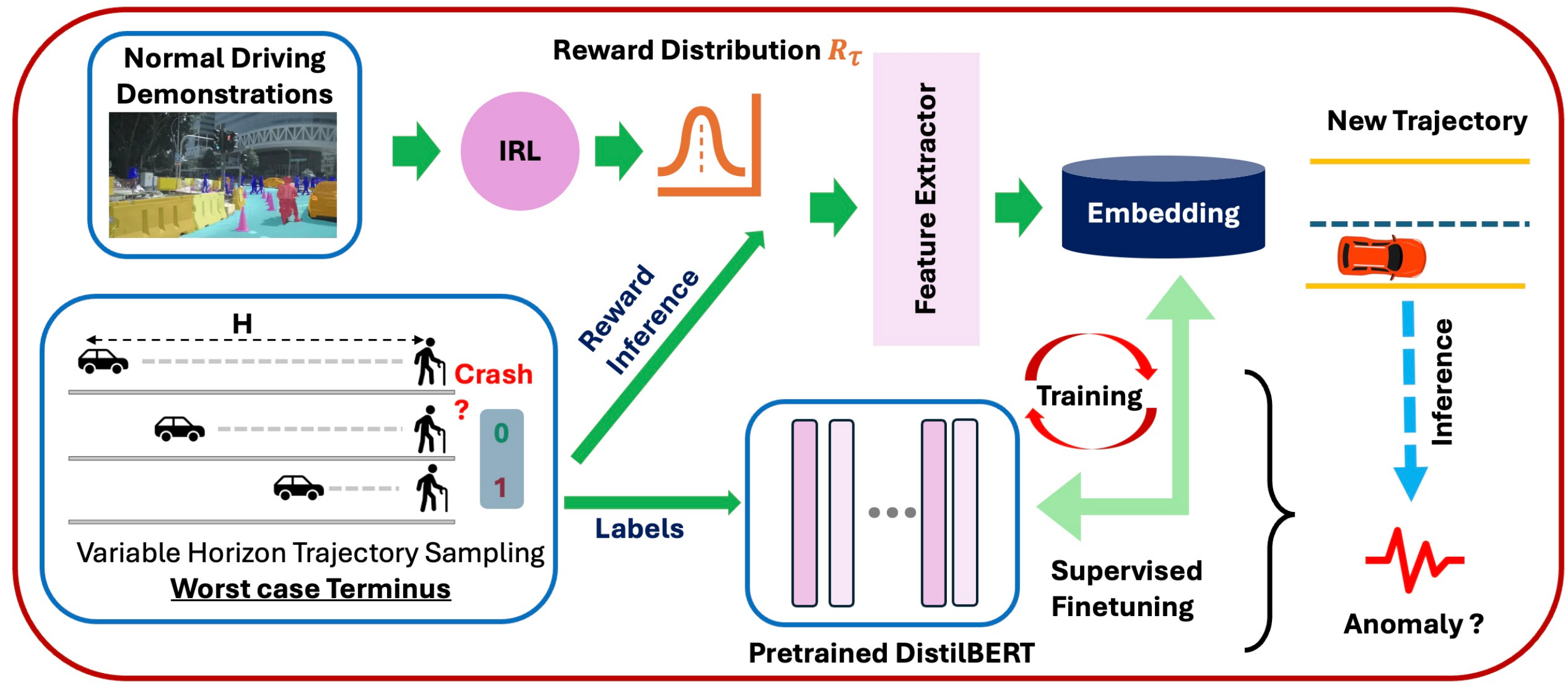}
        \caption{Hybrid IRL-based reward guidance in supervised fine-tuning of variable horizon sampled trajectories labeled for worst-case terminus. By making easy labeling of the worst-case trajectories, we can mostly avoid labeling for specific anomaly groups, which is costly. Our approach implicitly learns to classify even minor behavior deviations by maximizing the time to consequence.}
        \label{fig:architecture}
    \end{figure*}
    
\section{Problem Formulation}
    \subsection{Simulation environment}
    A flexible and efficient simulation environment is fundamental for reinforcement learning research, particularly when rapid prototyping and robust scenario customization are desired. The MetaDrive simulator \cite{li2022metadrive} offers extensive configurability of scenarios, observations, and action spaces.
     MetaDrive’s integration with ScenarioNet \cite{li2023scenarionet} enables recording and replaying trajectories while also being closed-loop simulation eligible. To improve training efficiency and resource utilization, we use vectorized environments, which enable parallel training across multiple instances—a capability that MetaDrive supports at scale. By processing actions, observations, rewards, and termination signals in batches across independent environments, the vectorized environment significantly accelerates RL training compared to the regular single-environment training paradigm.
        We modify MetaDrive for our custom settings, where each vehicle observes its surroundings using 240 LiDAR lasers (also referred to as LiDAR channels or pulses that measure distance to objects), as well as 12 lane detectors (detect broken/solid lines) and 12 side detectors (detect sidewalks/solid lines). We opt for LiDAR state observation over top-down or visual observation due to computational efficiency and storage constraints, as we can approximate our state with only a few lasers. 
        
        However, our trajectories are also separately stored in Scenario Description Language (SDL) as in ScenarioNet \cite{li2023scenarionet}, which lets us replay and derive the visual counterpart if required. Another major benefit of this simulator is the scenario customizability, which generates an almost unlimited number of scenarios using random sampling. Some of the most common ones in regard to popular road structures can be easily implemented. We can modify attributes such as lane width, number of vehicles, the gap of IDM-driven vehicles from the ego vehicle, and many more, thus leading to a new scenario every time we reset the environment. In our setup, each trajectory represents a unique scenario, created by randomizing the environment with a different random seed at every reset.

    \subsection{Rule based anomaly labeling}\label{subsec:rule-based}
       In this paper, we assume that the only given perception information is in the form of LiDAR, as well as lane and side detectors.  We make this assumption to accommodate for the fact that these sensors are available to most Level-3+ autonomous vehicles. Since the simulator also provides access to vehicle dynamics data, such as heading, acceleration etc.,  we utilize this additional information to generate trajectory anomaly targets through a set of rule-based labeling criteria (explained below) for evaluation purposes. However, during inference, we depend solely on perception data to determine whether a trajectory is anomalous. Below are the specific types of anomalies our labeling system detects:
       
    \subsubsection{Zigzag driving patterns}
    
    Given the vehicle heading $\theta_x$ and $\theta_y$, we detect the curvature of trajectory at any given time as follows:
    
     {\small
     \begin{equation}
         k=\frac{\left|\theta_x^{\prime} \theta_y^{\prime \prime}-\theta_y^{\prime} \theta_x^{\prime \prime}\right|}{\left(\theta_x^{\prime 2}+\theta_y^{\prime 2}\right)^{3 / 2}}
     \end{equation}}

    where $\theta_x^{\prime}=\frac{d \theta_x}{d t}, \theta_y^{\prime}=\frac{d \theta_y}{d t}, \theta_x^{\prime \prime}=\frac{d^2 \theta_x}{d t^2}, \theta_y^{\prime \prime}=\frac{d^2 \theta_y}{d t^2}$. We compute these time derivatives based on the temporal window size (40 steps in our case) and threshold this value based on human observation to detect high-frequency zig-zag patterns.
    \subsubsection{Sudden Braking}
    We calculate the velocity magnitude $V_i=\left\|\mathbf{v}_i\right\|$ for every time point $i$. Acceleration can thus be calculated using finite differences over a specified window size given by: 
    $A_i=\frac{S_{i+w / 2}-S_{i-w / 2}}{w}$ where $w$ is the window size and $A_i$ is the acceleration at point $i$. We then compute the smoothed acceleration using a moving average filter as $S A_i=\frac{1}{w} \sum_{k=-\frac{w-1}{2}}^{\frac{\mathrm{m}-1}{2}} A_{i+k}$. Here, the index $k$ runs between the half-windows $-\frac{w-1}{2}$ to $\frac{w-1}{2}$, which ensures that the window is centered at $i$ and includes $w$ terms. This formula assumes that $w$ is odd; if $w$ is even, you might need to adjust the limits accordingly or redefine the window slightly to maintain symmetry. Then we identify sudden acceleration changes when $\text{Indices}=\left\{i \mid S A_i<\theta\right\}$, where $\theta$ represents the braking threshold (which is typically a negative value indicating deceleration).
    
    \subsubsection{Sudden turns}
    Similarly, measuring lateral acceleration is the best way to detect sudden turns. Given a sequence of heading vectors $\mathbf{h}_i$ (where $i$ indexes the frame), the angular change $\Delta \theta_i$ between consecutive heading vectors can be calculated as: $\Delta \theta_i=\arccos \left(\frac{\mathbf{h}_{i-1} \cdot \mathbf{h}_i}{\left\|\mathbf{h}_{i-1}\right\|\left\|\mathbf{h}_i\right\|}\right)$, where $\mathbf{h}_{i-1}$ and $\mathbf{h}_i$ are normalized to unit vectors before computing the dot product. We clip values to ensure the argument of across remains within its valid domain of $[-1,1]$ to prevent undefined values due to floating-point precision issues. The lateral acceleration $a_{\text {lat }, i}$ based on the angular change and velocities is computed using:
    {\small
    $$
    a_{\mathrm{lat}, i}=\Delta \theta_i \times v_{i+1}
    $$}
    where $v_{i+1}$ is the magnitude of the velocity vector at frame $i+1$, and $\Delta \theta_i$ is the angular change from the previous frame. To identify frames where the lateral force exceeds a certain threshold $\tau$ (such as a critical lateral acceleration that might indicate a risk of skidding or rolling in vehicle dynamics), we use:
    $$
    \text { Indices }=\left\{i| | a_{\text {lat }, i} \mid>\tau\right\}
    $$

Here, $\tau$ is set to $0.8~\mathrm{m}/\mathrm{s}^2$, but this value can be adjusted based on empirical data or dynamic conditions as required. Our choice is informed by the findings of \cite{de2023standards}, where a 20\% discomfort reporting threshold corresponded to a maximum lateral acceleration of $0.8~\mathrm{m}/\mathrm{s}^2$.

    \subsubsection{Frequent lane switching/driving between two lanes}
    We detect this specific lane anomaly using lane break points. We trigger a boolean whenever a vehicle crosses a lane line and evaluate the crossing frequency to detect an anomalous vehicle trajectory. 
    The moving average $M$ of the data array $D$ is computed using a convolution operation. This operation involves sliding a window of size $w$ across $D$, and for each position, computing the average of the elements within the window given by:
    $M=D * \frac{1}{w} \mathbf{1}_w$, where, $\mathbf{1} _w$ is a vector of ones with length $w$, $*$ denotes the convolution operation and $\frac1w\mathbf{1}_w$ is the kernel used for the convolution, effectively averaging the elements in each window of size $w$. This is central to smoothing the data, reducing noise, and helping to identify sustained patterns indicative of lane changes.
    We then identify indices where $M$ exceeds 0.5 :
    $$
    \text { Indices }=\left\{i \mid M_i>0.5\right\}
    $$

    We derive contiguous index ranges as Intervals and then filter these by computing the length of each interval based on a threshold $\tau$. 
    {\small
    \begin{equation}
        \mathrm{Anomalies}=\{\ell(\mathrm{Int.})\mid\mathrm{Int.}\in\mathrm{Intervals~}\wedge\ell(\mathrm{Int.})>\tau\}
    \end{equation}}
Here, $\ell(\mathrm{Int.})$ is the length of a specific interval and $\tau$ is the user-validated threshold indicating a significant duration indicative of an anomaly in lane change behavior.

    \subsubsection{Tailgating/Driving very close}
    We use the simulator-provided proximity detectors to detect if any vehicles exist close to a specific threshold to detect these behaviors. Thus, we directly label trajectories for these categories with pure vehicle proximity bounds.

\section{Algorithm Design}
\subsection{Reward Learning}
IRL aims at identifying reward estimates that explain expert behaviors and reason about their specific actions within the seen trajectories. Maximum Entropy IRL  \cite{ziebart2008maximum, finn2016guided} has been the most popular way of modeling reward estimates. With Maximum Entropy IRL, we aim to learn the underlying reward function that explains expert behavior by assuming experts are exponentially more likely to choose higher-reward actions, but not always perfectly. It does this by finding the reward function that makes the expert’s actions most probable while also maximizing randomness (entropy), so it doesn’t assume any more than what the data shows. In other words, this can be explained using a Boltzmann distribution as follows:

\begin{equation}
    p(\tau|\theta)=\frac{1}{Z}\exp(R(\tau|\theta))
\end{equation}

However, one major limitation of this approach is the assumption that the expert is optimal, which does not apply to most cases, even for well-trained policies or IDM (Intelligent Driver Model) policies. The notion of optimality indirectly biases the learned reward estimate, which results in even poor performance when a new policy is trained with this reward estimate, or simply detecting anomalies based on this reward, flaws the binary estimate. We thus implement the trajectory-ranked reward extrapolation approach \cite{brown2019extrapolating} to derive the reward function. With this approach, we first position the expert policy with varying noise intensity and then rank the expert rollouts $\tau_1,\ldots,\tau_m$, from worst to best. Thus with a reward estimate $\hat{r}_{\theta}(s)$ parameterized by $\theta$, we have $\sum_{s\in\tau_i}\hat{r}_\theta(s)<\sum_{s\in\tau_j}\hat{r}_\theta(s)$ when $\tau_i\prec\tau_j$. The reward function can then be trained in a supervised fashion with ranked demonstrations using the generalized loss function below:

\begin{equation}
    \label{eq:irl_loss}
    \mathcal{L}(\theta)=\mathbf{E}_{\tau_i,\tau_j\Pi}\left[\xi\left(\mathbf{P}(\hat{J}_\theta(\tau_i)<\hat{J}_\theta(\tau_j)),\tau_i\prec\tau_j\right)\right]
\end{equation}

where, $\Pi$ is a distribution over demonstrations, $\prec$ is the preference ordering between trajectories, $\xi$ is the binary classification loss function and $\hat{J}$ is a discounted return defined by a parameterized reward function $\hat{r}_{\theta}$, i.e. $\hat{J}_\theta(\tau) =\sum_{s\in\tau}\hat{r}_\theta(s)$

The probability $\mathbf{P}$ signifies this hierarchy between the trajectories and can be obtained using a softmax-normalized distribution. 

\begin{equation}
   \mathbf{P}\left(\hat{J}_\theta(\tau_i)<\hat{J}_\theta(\tau_j)\right)\approx\frac{\exp\sum_{s\in\tau_j}\hat{r}_\theta(s)}{\exp\sum_{s\in\tau_i}\hat{r}_\theta(s)+\exp\sum_{s\in\tau_j}\hat{r}_\theta(s)}, 
\end{equation}

This objective is trained by predicting whether one trajectory is preferable to another, with the label being the hierarchical ordering that we already know based on the added noise to expert policy. The output of this training is the learned reward function $\hat{r}_{\theta}(s)$.

\subsection{Worst-Case-Terminus}
    Defining an anomaly is a significant challenge, specifically in autonomous driving scenarios, due to dynamic factors such as environmental conditions, road user behavior, sensor limitations, etc. A more reasonable way to define an anomaly is the deviation that leads to potential hazards or system failures. Anomalous behaviors that do not lead to any hazard are of significantly low interest, especially in safety-critical scenarios such as autonomous driving. Thus, we define worst-case terminus as the utmost severe consequence that a vehicle might have to suffer if it behaves abnormally. In our case, we define vehicles crashing with objects (such as traffic cones, walls, other vehicles etc.) and driving off the road leading to simulation termination as worst-case terminus.
    One benefit of defining worst-case-terminus is that these are limited, unlike anomalies, which can appear in different forms every single time and are explicitly hard to dictate. This helps us generate binary labels if the worst-case terminus is reached for each expert trajectory, essentially turning into a supervised learning setting. In our implementation, we thus generate these crashes by different ways of traffic density variation and adding noise to the expert policy.
    
\begin{figure}[t]
    \centering
    \includegraphics[width=0.8\linewidth]{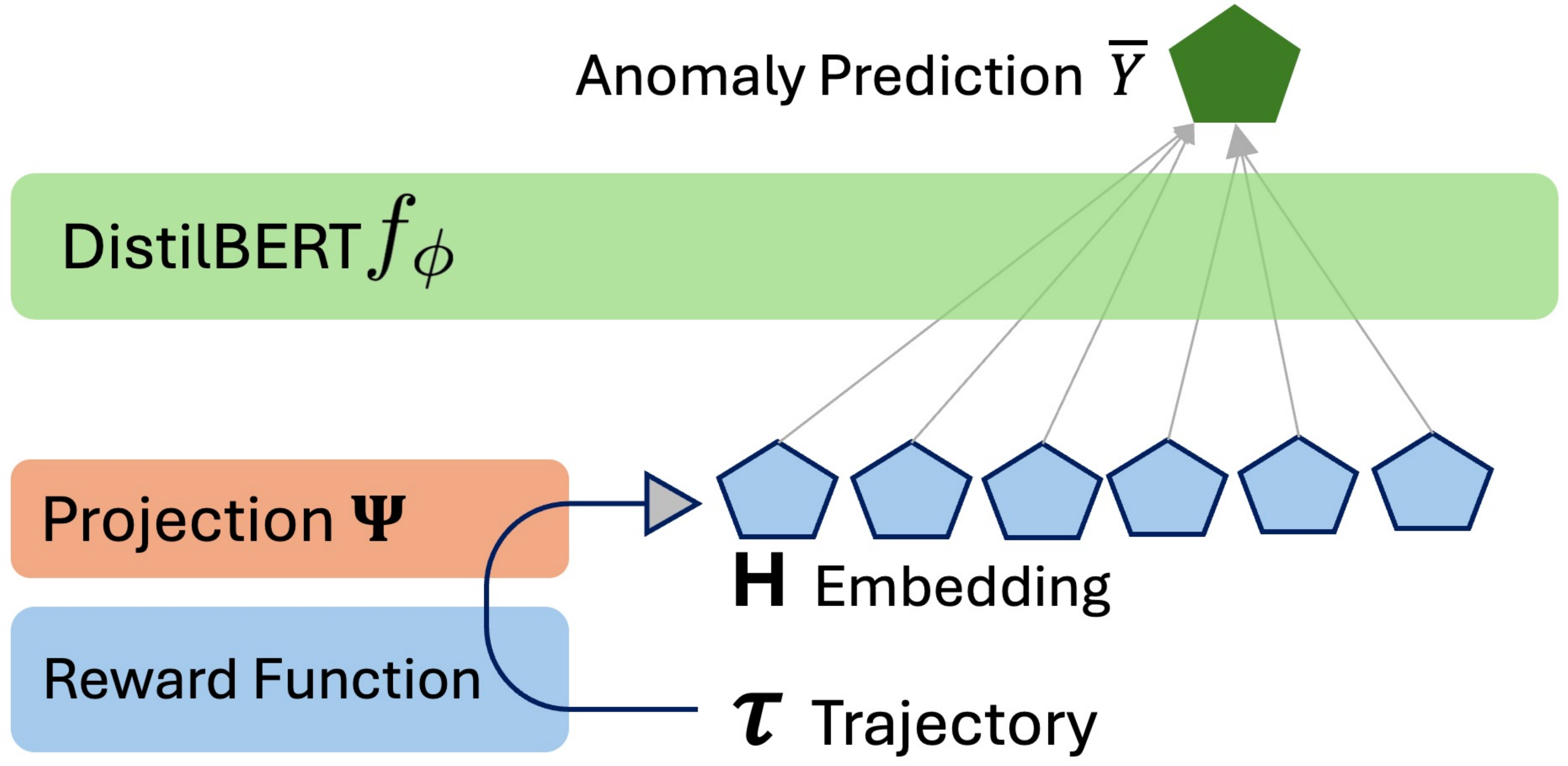}
    \caption{Trajectory-Reward fusion to generate DistilBERT compatible embedding for binary classification }
    \label{fig:enter-label}
\end{figure}

\begin{figure*}[ht!]
    \centering
    \includegraphics[width=0.9\linewidth]{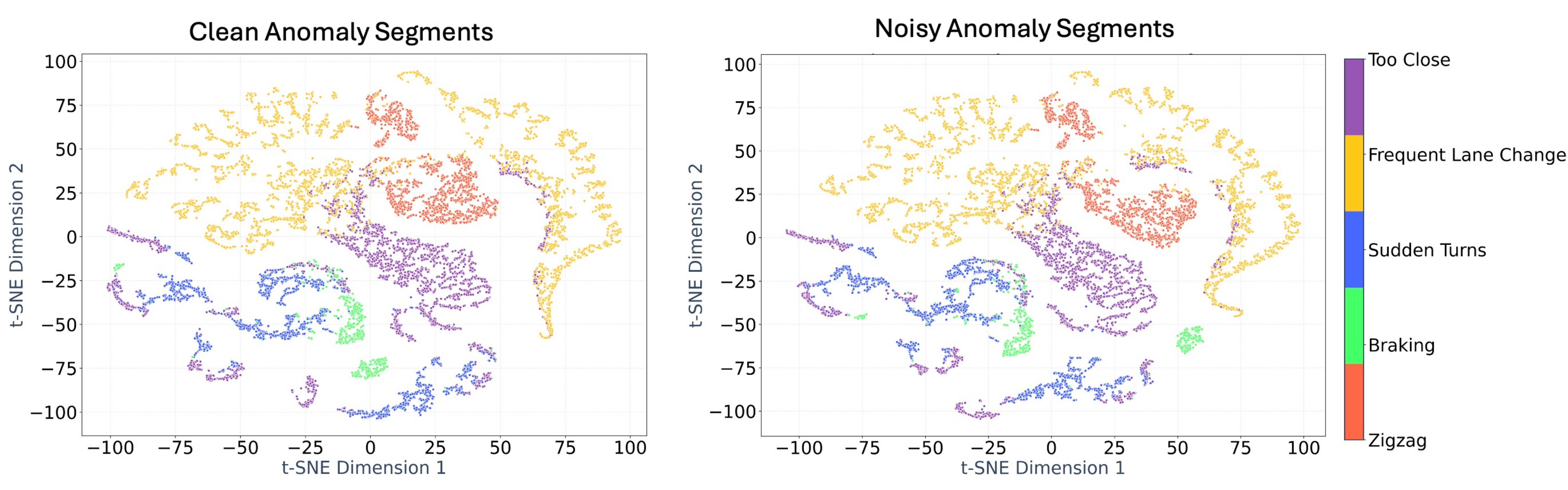}
    \caption{t-SNE plot of anomalous segments of test trajectories labeled using the rule-based approach described in Section \ref{subsec:rule-based}. Figure shows distribution shift with added noise(type: Composite2 high) }
    \label{fig:t-sne-segment}
\end{figure*}

    \begin{figure}[h!]
        \centering
        \includegraphics[width=0.8\linewidth]{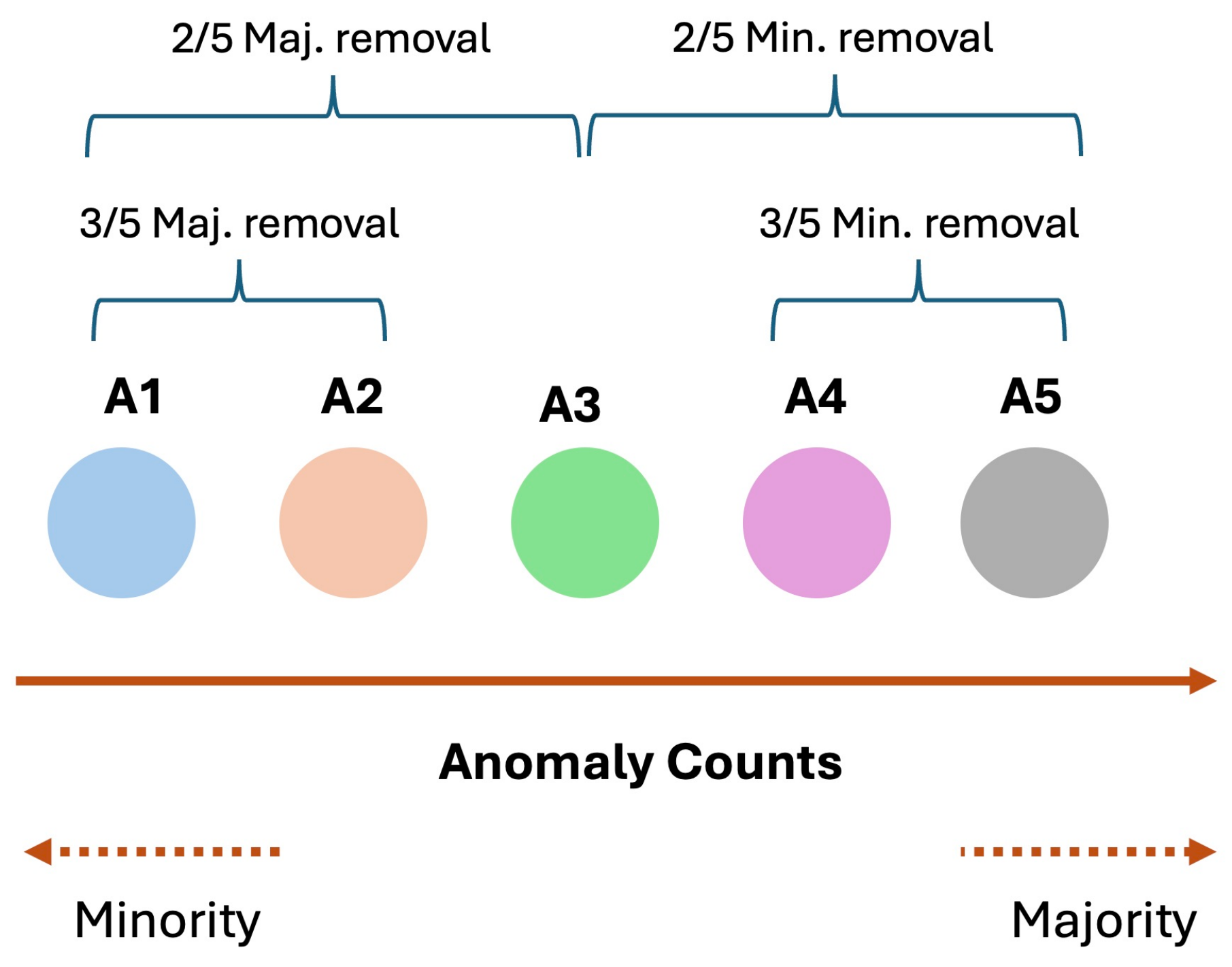}
        \caption{Figure demonstrating the expert anomaly composition formation. We form four different composition groups by removing trajectories corresponding to the increasing majority and minority groups. }
        \label{fig:anomaly_composition}
    \end{figure}
    
\subsection{End-Clipping \& Variable Trajectory Sampling}
    Defining the worst-case terminus simply does not allow us to detect anomalies, and training for crash and no-crash on its own is a trivial problem to solve, considering the early deviation detection objective. However, we take a different approach and intend to maximize the time for the worst-case-terminus. This induces the side effect of detecting the change from normal to abnormal behavior that leads to severe consequences. This is implicitly achieved with variable sub-trajectory sampling, where we break demonstrations into different trajectories of high variance in lengths. This strategy needs to be applied at the dataset level. Let $D=\{\tau_i\}_{i=1}^N$ be the training trajectory datasets containing worst-case-terminus and let each trajectory have length $L_{i}$. The max-length and min-length of trajectories in the dataset are represented by $L_{\max}=\max_{i\in D}L_i$ and $L_{\min}=\min_{i\in D}L_i$. The trajectory sampling steps are as follows:
    \newline
    \textbf{Step 1:} Crop each trajectory from the end with a random length
        $L_{rand}\sim\mathrm{Uniform}(\min(L_{min},L_{thres}),L_{i})$, where $L_{thres}$ is the max length the cropped trajectories can have.
    \newline
    \textbf{Step 2:} Expand each trajectory multiple times.
    Let each trajectory be defined by $\tau=(o,a,r,l)$ where o is the observation, a is the action, $r$ is the reward, and $l$ is the binary label for the worst-case-terminus. Note $r$ here is obtained using reward inference of the state observation $o$ from the learned reward estimate $\hat{r}_{\theta}$.
    Let $\mathcal{L}$ be the set of $\eta$ integers sampled uniformly at random from the range [2,n) such that $\mathcal{L}\subseteq\{2,3,\ldots,n-1\},\quad|\mathcal{L}|=\min(\eta,n-2)$. Then for each $k\in\mathcal{L}$, we expand the trajectory as $$\tau_k^{\prime}=(o[:k],a[:k],r[:k],l).$$ 
    We do this for every trajectory with worst-case-terminus or with label = 1 as follows:

    \begin{equation}
        D^{\prime}=\bigcup_{\tau\in D; l=1}\text{expand traj}(\tau)
    \end{equation}

\subsection{Finetuning DistilBERT}
We use DistilBERT \cite{sanh2019distilbert}, which is a compact and efficient version of BERT (Bidirectional Encoder Representations from Transformers) \cite{devlin2019bert} language model, as our anomaly detection module for sequential classification. Given the specially configured trajectories from our new dataset $D^{\prime}$, we generate embeddings for each timestep corresponding to these trajectories. Note that these trajectories already have reward values estimated from our reward function $\hat{r}_{\theta}$, which is learned using IRL through the minimization of equation (\ref{eq:irl_loss}). 
We now define a projection feedforward network $\Psi$ that takes state, action, and reward estimate and generates an embedding token $h$, i.e., is compatible with DistilBERT's embedding dimension. We finally finetune DistilBERT to classify each trajectory of embedding token $h$ for the corresponding label $l$.
\begin{equation}
    \textbf{h} = \Psi(o, a, r)
\end{equation}

    Thus the final training dataset $D^{\prime}$ is given by $D^{\prime} = D \cup D^{\prime}$. Note that this trick directly contributes to optimization for time maximization and is not a part of dataset generation by itself.

    Note, during inference, the test trajectories may or may not include crash points. Naturally, our model is likely to flag these crash points as anomalies, which is expected behavior. However, since our training objective is to detect such worst-case endpoints, including them during evaluation can introduce bias. To ensure a fair assessment of the model’s performance, we trim the final steps of each trajectory, thereby removing this bias and allowing for a more balanced evaluation.

\begin{figure}[t]
    \centering
    \includegraphics[width=0.8\linewidth]{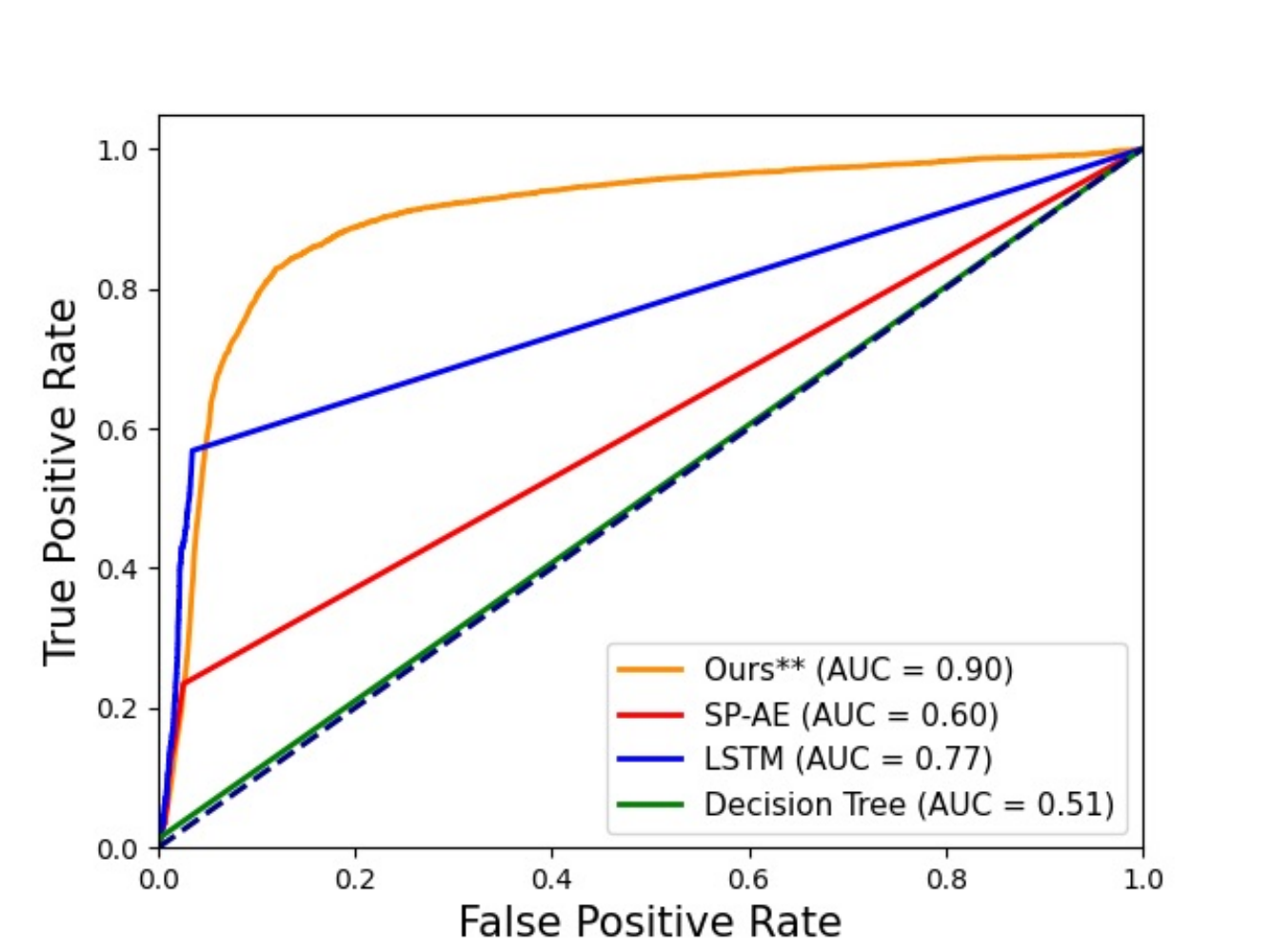}
    \caption{ROC curve demonstrating the performance comparison of some widely used anomaly detection methods trained with WCT.}
    \label{fig:enter-label}
\end{figure}

\subsection{Implicit optimization strategy}
Our method is related to Receding Horizon Optimization (RHO) \cite{mayne1988receding}, also known as Model Predictive Control (MPC), which is a feedback control strategy that repeatedly solves a finite-horizon optimization problem through online learning. An optimal control sequence is computed over a prediction horizon after the first control action is applied, which shifts the horizon step forward. One popular way to deal with piecewise continuous nominal control trajectory in such problems is using variable horizon length in each iteration \cite{bergman2020optimization, vahidi2024efficient}, which aligns with our strategy. Another analogy that includes RHO is the state estimation whereas our approach implements reward-guided embedding projection. However, the core optimization in our implementation occurs through temporal reward decomposition and variable-length trajectory embedding. TRAP learns temporal relationships through data-driven reward shaping rather than explicit trajectory optimization. This makes it more adaptable to unknown anomaly patterns while maintaining computational efficiency. Specifically, our approach implicitly solves three interconnected optimization processes:
    \subsubsection{Temporal Credit Assignment}
    When cropping trajectories to random lengths, we essentially solve the following:
        \small
        \begin{equation}
        \max_\theta \; \mathbb{E}_{L_i, \tau \sim D'} \bigg[ \sum_{t=1}^{L_i} \gamma^t \hat{\tau}_\theta(s_t) + \lambda \cdot \mathrm{KL}\left( p(\tau \mid \theta) \,\|\, p_{\text{expert}}(\tau) \right) \bigg]
        \end{equation}
    where, $L_i$ is a random trajectory length and \small $L_{i}\sim \mathcal{U}[L_{\min},  L_{\text{thres}}]$ with both $L_{min}$ and $L_{thres}$ being discrete valued.
    This forces the reward function $\hat{r}_\theta$ to:
    \begin{itemize}
        \item Prioritize early warning signs in short segments. Since $L_i$ is randomly sampled and can be small, the model must extract information from early states and weight accordingly to incentivize early timesteps. 
        \item Maintain temporal consistency across partial observations. To generalize across random-length segments, $r_\theta$ must be weighted to behave consistently in time, even with partial observations. This is further strengthen by the possibility of overlap between these segments and reward structure must match regardless.
        \item Balance immediate rewards against long-term consequences. Much like the weighting effect of $\gamma^t$, our approach through whole sequence modeling learns to tradeoff short and long term behavior. 
    \end{itemize}
    The KL term penalizes divergence from the expert distribution which further anchors the trajectory level structure and prevents from overfitting to short-term patterns if those cause long-term divergence from expert thus maintaining balance among all three steps. This in our case is attained by the minimizing the cross-entropy loss.

    \subsubsection{Worst-case anticipation and implicit horizon coupling}
    The trajectory clipping creates an adversarial optimization landscape: 
    
    {\small
    \begin{equation}
        \min _\phi \mathbb{E}_{r \in D^{\prime}}\left[1_{\mathrm{WCT}} \cdot \exp \left(-\alpha \sum_{t=1}^k \hat{r}_\theta\left(s_t\right)\right)\right]
    \end{equation}}
    where $\phi$ are the DistilBERT parameters. This penalizes the model exponentially when high-reward trajectories (reward scaled) lead to crashes (false negatives) or low-reward trajectories avoid crashes (false positives).

    Moreover, the variable lengths paired with clipping create an implicit relationship between detection time $t_d$ and the remaining time to terminus $T_{\text{WCT}}$ given by:
    {\small
    \begin{equation}
       \mathrm{P}\left(\mathrm{WCT} \mid \tau_{1:t_d}\right) \propto \exp \left(-\beta \frac{T_{\mathrm{WCT}}}{t_d} \sum_{k=1}^{t_d} \hat{r}_\theta\left(s_k\right)\right)
    \end{equation}}
    This automatically weights early detections more heavily when $T_{\text{WCT}}$ is small.

    \subsubsection{Multi-scale feature learning}
        Variable lengths essentially induce a curriculum learning effect by asking if the trajectory segment (ahead in time) anticipates the future worst case: 
        \small
        \begin{equation}
            \mathcal{L}_{\mathrm{emb}}=\sum_{k=1}^K w_k \cdot \operatorname{CE}\left(\Psi\left(o_{: k}, a_{: k}, r_{: k}\right), l_{\mathrm{WCT}}\right)
        \end{equation}
        where weights $w_k$ increase with segment length $k$ and CE is the cross entropy. This makes the model focus on local patterns for early detection of deviations, integrates global context for final classification and learns length-invariant representations.

\section{Experiments}
We compare both supervised and unsupervised methods in our evaluation. Our method is supervised for the worst-case terminus but remains unsupervised for anomaly detection. 
For the purpose of our evaluation, we consider three of the most popular yet powerful unsupervised anomaly detection models, i.e., One-Class SVM (OC-SVM) \cite{manevitz2001one}, Isolation Forest (IForest) \cite{liu2008isolation}, and Local Outlier Factor (LOF) \cite{breunig2000lof}, and three powerful supervised classifiers, namely Long short-term memory (LSTM) \cite{hochreiter1997long}, Decision Tree \cite{quinlan1986induction}, Supervised Autoencoder (SPAE) \cite{le2018supervised}. We also consider the popular IRL based Sequential Anomaly Detection algorithm by Oh \textit{et al.} \cite{oh2019sequential} (Seq-AD). For a fair comparison, since worst-case terminus and reward modeling are trained on expert scenarios, we train supervised methods on expert trajectories for the worst-case terminus and evaluate them on the same test dataset. Similarly, we train unsupervised methods on expert trajectories, which, despite being imperfect and containing anomalies, provide sufficient data for learning the underlying anomaly structure. We utilized 5,000 expert trajectories to train our model. For non-sequential methods, including both supervised and unsupervised approaches such as Decision Trees and SPAE, we concatenated and flattened state-action data to generate compatible inputs. Given the intensive computational requirements of some unsupervised methods (e.g., OC-SVM), processing large datasets is challenging. However, we found that using more than 5,000 trajectories did not significantly improve results. For a more comprehensive assessment, we conducted an evaluation of 14,000+ trajectories. These trajectories have the following characteristics:

\begin{itemize}
    \item  Can be \textbf{replayed and simulated} in closed form
    \item  Include \textbf{anomaly labeling} for each trajectory
\end{itemize}

This expanded dataset allows for a more robust evaluation of our model's performance across a diverse set of scenarios.

\begin{figure*}
    \centering
    \includegraphics[width=0.9\linewidth]{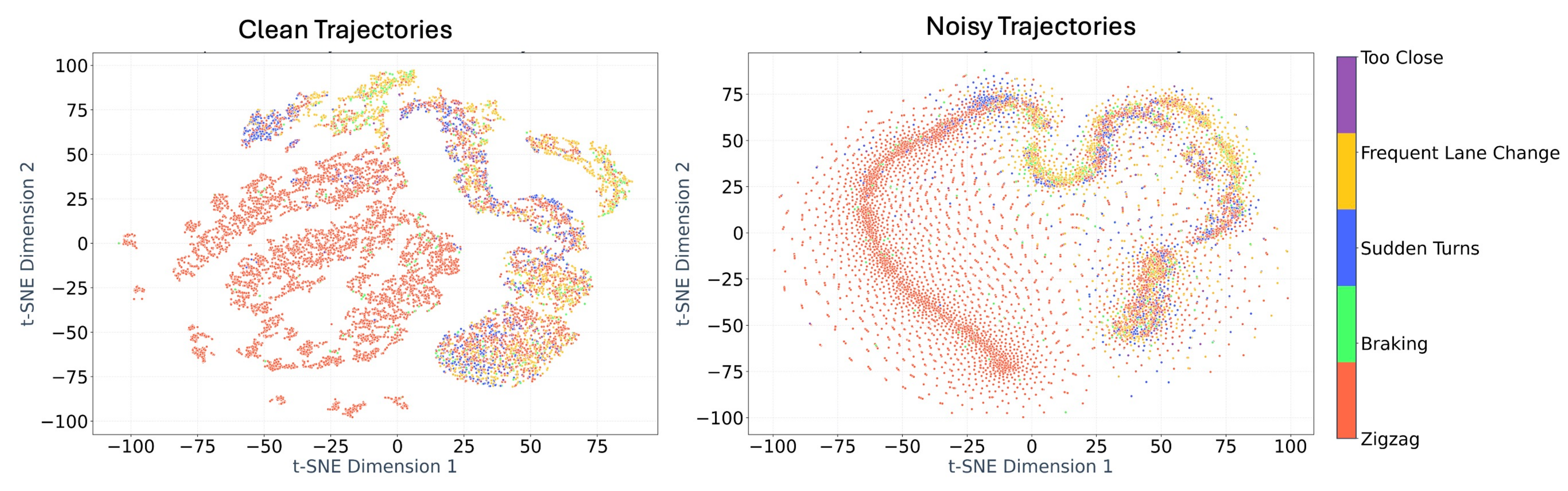}
    \caption{t-SNE plot of 14000+ full test trajectories labeled using the rule-based approach described in \ref{subsec:rule-based} for evaluation. Figure shows distribution shift with added noise (type: Composite2 high)}
    \label{fig:t-sne-full}
\end{figure*}

\begin{table*}[htpb!]
\centering
\caption{Comparison of TRAP with various Anomaly Detection methods}
\label{tab:normal_comparision}
\resizebox{0.8\textwidth}{!}{%
\begin{tabular}{@{}lllllll}
\toprule
\textbf{Models} & \textbf{F1-Score} & \textbf{AUC} & \textbf{Precision} & \textbf{Recall} & \textbf{MCC} & \textbf{Accuracy} \\ \midrule
OCSVM              & 0.611& -& 0.503& 0.777& 0.185& 0.564\\
IForest               & 0.688& - & \textbf{0.949}& 0.539& 0.592& 0.784\\
LOF               & 0.498& -& 0.724& 0.379& 0.313& 0.663\\
\hline
LSTM (Trained WCT) & 0.656& 0.766& 0.925& 0.508& 0.552& 0.765\\
Decision Tree (Trained WCT)               & 0.029& 0.51& 0.838& 0.015& 0.071& 0.565\\
SPAE (Trained WCT)               & 0.37& 0.604& 0.878& 0.234& 0.322& 0.649\\
\hline
Seq-AD (Final)                & 0.411& -& 0.318& 0.579& -0.491& 0.269\\
\hline
\textbf{Ours (Final)**}            & \textbf{0.822}& \textbf{0.90}& 0.754& \textbf{0.902}& \textbf{0.667}& \textbf{0.828}\\ \bottomrule
\end{tabular}%
}
\end{table*}

\begin{figure}[ht!]
    \centering
    \includegraphics[width=\linewidth]{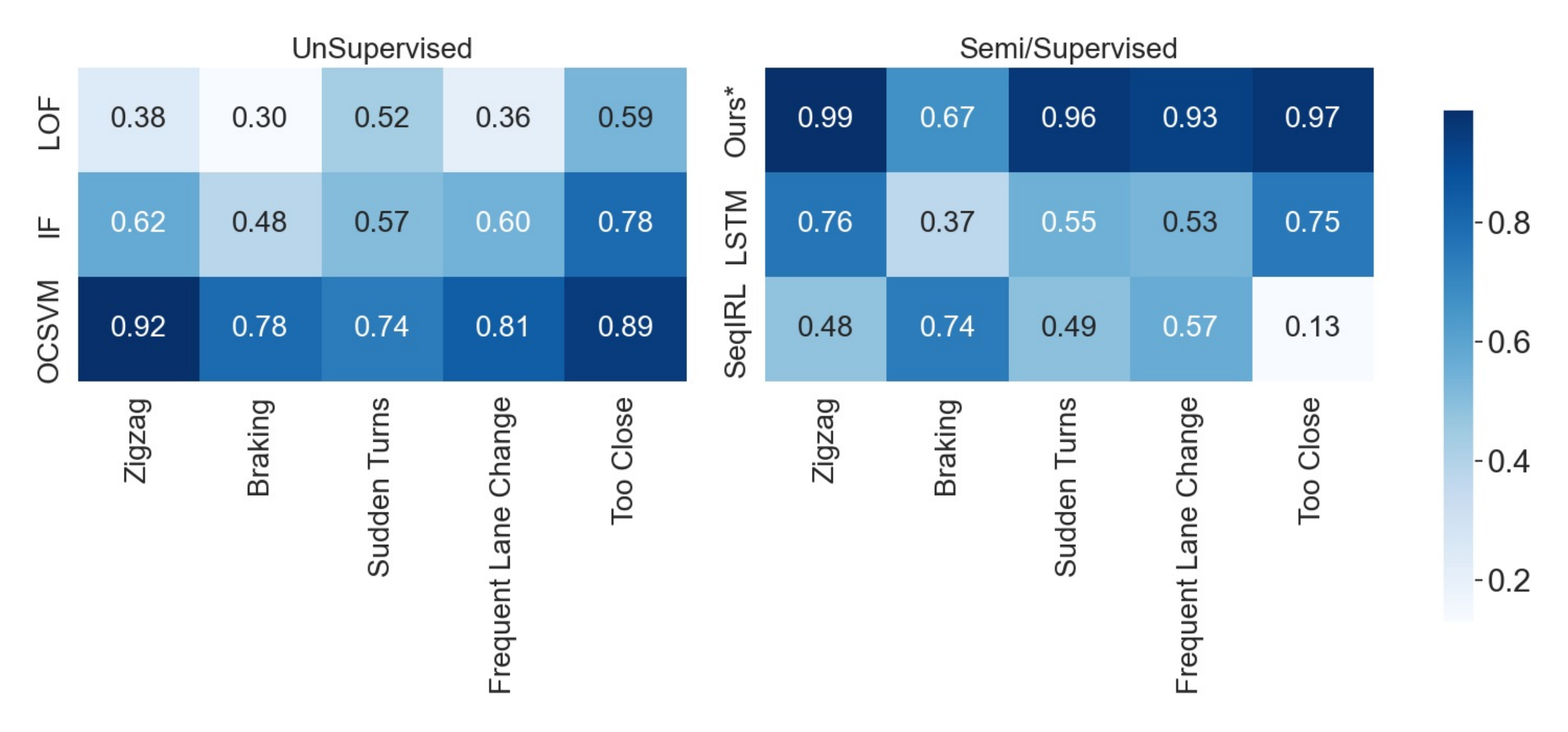}
    \caption{Figure shows the percentage of anomalies identified by unsupervised (on the left) and semi-supervised methods (on the right) on test trajectories.}
    \label{fig:confusion_plot}
\end{figure}

\begin{table*}[ht]
\centering
\caption{Performance Comparison: F1 Scores and Mathews Correlation Coefficient (MCC)}

\resizebox{0.9\textwidth}{!}{%
\begin{tabular}{lccccccccccccccccc}
\toprule
\textbf{Noise Type} & \textbf{Metric} & & \multicolumn{3}{|c|}{\textbf{OCSVM}} & \multicolumn{3}{|c|}{\textbf{Isolation Forest}} & \multicolumn{3}{|c|}{\textbf{LSTM}} & \multicolumn{3}{|c|}{\textbf{SPAE}} & \multicolumn{3}{|c|}{\textbf{Ours*}} \\
                    &  & & \textbf{Low} & \textbf{Med} & \textbf{High} & \textbf{Low} & \textbf{Med} & \textbf{High} & \textbf{Low} & \textbf{Med} & \textbf{High} & \textbf{Low} & \textbf{Med} & \textbf{High} & \textbf{Low} & \textbf{Med} & \textbf{High} \\
\midrule
\rowcolor{lightgray}
\textbf{Precipitation} & F1  & & 0.496 & 0.597 & 0.603 & 0.693 & 0.710 & 0.747 & 0.611 & 0.611 & 0.611 & 0.050 & 0.050 & 0.050 & \textbf{0.824} & \textbf{0.826} & \textbf{0.831} \\
\rowcolor{lightgray}
                         & MCC & & 0.174 & 0.041 & -0.075 & 0.596 & 0.610 & 0.640 & 0.515 & 0.515 & 0.515 & 0.094 & 0.094 & 0.094 & \textbf{0.672} & \textbf{0.677} & \textbf{0.688} \\
\addlinespace
\textbf{Distance}       & F1  & & 0.596 & 0.595 & 0.594 & 0.707 & 0.711 & 0.716 & 0.611 & 0.611 & 0.611 & 0.051 & 0.051 & 0.051 & \textbf{0.822} & \textbf{0.822} & \textbf{0.822} \\
                         & MCC & & 0.161 & 0.163 & 0.165 & 0.608 & 0.611 & 0.616 & 0.515 & 0.515 & 0.515 & 0.095 & 0.095 & 0.095 & \textbf{0.667} & \textbf{0.667} & \textbf{0.667} \\
\addlinespace
\rowcolor{lightgray}
\textbf{Ambient}        & F1  & & 0.611 & 0.611 & 0.611 & 0.781 & \textbf{0.834} & 0.617 & 0.638 & 0.714 & 0.626 & 0.049 & 0.038 & 0.004 & \textbf{0.819} & 0.814 & \textbf{0.803} \\
\rowcolor{lightgray}
                         & MCC & & 0.000 & 0.000 & 0.000 & \textbf{0.669} & \textbf{0.696} & 0.101 & 0.537 & 0.607 & 0.414 & 0.094 & 0.080 & 0.031 & 0.661 & 0.652 & \textbf{0.630} \\
\addlinespace
\textbf{Spurious}       & F1  & & 0.611 & 0.592 & 0.611 & 0.698 & 0.750 & 0.813 & 0.611 & 0.611 & 0.610 & 0.050 & 0.050 & 0.051 & \textbf{0.822} & \textbf{0.824} & \textbf{0.829} \\
                         & MCC & & 0.178 & -0.067 & -0.026 & 0.599 & 0.645 & \textbf{0.704} & 0.515 & 0.515 & 0.515 & 0.094 & 0.094 & 0.095 & \textbf{0.668} & \textbf{0.672} & 0.681 \\
\addlinespace
\rowcolor{lightgray}
\textbf{Composite1}       & F1  & &  0.611&             0.611&             0.611&             0.802&             \textbf{0.825}&              0.612&0.638             &0.714             &0.636              &0.049             &0.036             &0.004                         & \textbf{0.820} & 0.821 & \textbf{0.824}\\
\rowcolor{lightgray}
                         & MCC & &  0.0&             0.0&             0.0&0.690             &\textbf{0.674}             &0.042              &0.537             &0.606             &0.417              &0.094             &0.078             &0.029                                   & \textbf{0.663} & 0.666 & \textbf{0.671}\\
\addlinespace
\textbf{Gaussian}       & F1  & & 0.599&             0.605&             0.619&             0.752&             0.782&              0.819&0.612             &0.613             &0.618              &0.051             &0.050             &0.050               &\textbf{0.822 }& \textbf{0.822}& \textbf{0.822}\\
                         & MCC & & 0.199&             0.233&             0.281&0.646             &\textbf{0.672}             &\textbf{0.710}              &0.516             &0.517             &0.521              &0.095             &0.094             &0.094                              & \textbf{0.667} & 0.667 & 0.667\\
\addlinespace
\rowcolor{lightgray}
\textbf{Dropout}       & F1  & & 0.611&             0.611&             0.611&             0.845&             0.865&              0.873&0.611             &0.611             &0.611              &0.051             &0.050             &0.051                & \textbf{0.856 }& \textbf{0.873 }& \textbf{0.877}\\
\rowcolor{lightgray}
                         & MCC & & 0.0&             0.0&             0.0&0.741             &0.768             &0.778              &0.515             &0.515             &0.515              &0.095             &0.094             &0.095                                     & \textbf{0.739} & \textbf{0.773} & \textbf{0.782}\\
\addlinespace
\textbf{Composite2}       & F1  & &  0.611&             0.611&             0.611&             0.859&             \textbf{0.874}&              0.853&0.612             &0.613             &0.615              &0.051             &0.049             &0.050             & \textbf{0.859} & 0.872 & \textbf{0.875}\\
                         & MCC & &  0.0&             0.0&             0.0&\textbf{0.759}             &\textbf{0.779}             &0.729              &0.516             &0.517             &0.518              &0.095             &0.093             &0.093                                   & 0.744 & 0.771 & \textbf{0.778}\\
\addlinespace
\midrule
\end{tabular}
}
\end{table*}
 
    \subsection{Noise Robustness}
        We introduce eight novel noise groups designed to simulate real-world conditions \cite{smith_lidar_modelling} affecting LiDAR sensor readings. These noise types are applied to the measurements of 240 LiDAR sensors in various forms:

\begin{itemize}
    \item \underline{Precipitation} noise simulates rain-induced light refraction and time delay \cite{haider2023methodology, goodin2019predicting}.
    \item \underline{Distance} noise represents inaccuracies in distance measurements due to sensor errors \cite{espadinha2021lidar}.
    \item \underline{Ambient} noise emulates environmental factors such as fog, sunlight, dust, and humidity \cite{2018DLPD}.
    \item \underline{Gaussian} noise introduces random noise affecting the entire beam uniformly \cite{haider2023methodology}.
    \item \underline{Dropout} noise models sensor malfunctions by randomly zeroing out measurements \cite{espadinha2021lidar, nakashima2021learning}.
\end{itemize}
        Since these noise types are applied individually to each LiDAR beam, we maintain precise control over their impact on the overall sensor readings. Additionally, we define two composite noise configurations that combine fundamental noise types in distinct ways:

\begin{itemize}
    \item \underline{Composite1}: A combination of precipitation, distance, ambient, and spurious noise.
    \item \underline{Composite2}: A mix of Gaussian, dropout, and spurious noise.
\end{itemize}
These noise models enable a more comprehensive evaluation of LiDAR sensor robustness under real-world perturbations.
        
        We note the rather unusual behavior of increased accuracy with increased noise which is not uncommon in unsupervised evaluation. This behavior is also observed in other models under comparison. The proposed results indicate the following:
            \begin{itemize}
                \item Our method proves to be inherently robust to certain types, thus letting us perform better by ignoring some inherent irrelevant variations.
                \item Noise, in many cases, can enhance separation and the contrast between normal and anomalous patterns, making anomalies easier to detect. This increased distribution separation can also be visualized in Figure \ref{fig:t-sne-full} on the right (noisy) compared to the normal on the left. 
            \end{itemize}

    Another key point to note is that recall is more valuable when missing anomalies are costly (e.g., missing to detect a vehicle driving in the wrong lane). Precision is more valuable when false alarms are costly (e.g., making too many false predictions might induce consequences). In ours and most anomaly detection scenarios, \textbf{high recall is prioritized} since missing anomalies can have severe consequences. However, a balance with precision is also needed to avoid too many false positives. In our evaluation, as shown in Table \ref{tab:normal_comparision}, we can observe that our method achieves significantly higher recall compared to all the other methods while also maintaining reasonable precision. Isolation forest, even having higher F1 compared to other methods, seems to have significantly lower recall, which explains that it misses a significantly larger proportion of anomalous events. Isolation Forest, despite having a higher F1 score than other methods, exhibits significantly lower recall, indicating it misses a larger proportion of anomalous events.

    \subsection{Normal Trajectory Evaluation without additional noise}

        This is the most basic mixed training configuration, where no consideration is given to the types of anomalies present in the expert trajectories. Additionally, no extra noise is introduced, and evaluation is performed in its raw form. The key difference between expert and test trajectories here lies in their sampling from different seed ranges, leading to variations in distribution.

        As shown in Table \ref{tab:normal_comparision}, TRAP demonstrates a significant performance advantage, achieving an F1 score improvement of +0.134. Isolation Forest, despite its high precision, only makes correct predictions when highly confident. However, its low recall suggests it misses a large number of true positive cases, which is costly.

        Furthermore, Sequential IRL performed poorly, often making predictions opposite to the expected outcome, as indicated by its MCC score. Decision Tree exhibited near-random performance, yielding the lowest F1 score overall. In contrast, LSTM outperformed all other supervised approaches, achieving the highest AUC among supervised models, indicating its effectiveness in this setting.

        Figure \ref{fig:confusion_plot} illustrates the percentage of anomalies detected across different anomaly groups in the test trajectories. Seq-AD performs well in the braking scenario but fails in nearly all other cases. Notably, despite applying a grid search to optimize the threshold for maximum F1 score, Seq-AD still under-performs compared to other methods. Similarly, OCSVM achieves high performance in the braking scenario, even surpassing TRAP, but performs worse than our approach in all other cases. Overall, TRAP demonstrates a significant performance advantage across most scenarios, highlighting its robustness in anomaly detection.

    \subsection{Expert trajectories biased to specific anomaly types}
    We perform additional experiments to evaluate how trajectories pertaining to certain anomalies might influence the predictions on the test trajectories, as shown in Table \ref{tab:group_removal}.
        We consider another approach where an expert dataset might include certain types of anomalies while missing out on some others. We consider 4 different approaches to account for this:
        \begin{itemize}
            \small
            \item Train by omitting \textbf{2 majority} anomaly groups from the expert
            \item Train by omitting \textbf{3 majority} anomaly groups from the expert
            \item Train by omitting \textbf{2 minority} anomaly groups from the expert
            \item Train by omitting \textbf{3 minority} anomaly groups from the expert
        \end{itemize}
        This is more practical in the real world, where normal driving behaviors might not include some of the serious or even menial anomaly types, while others are more apparent. For instance, braking and sudden turns might be more apparent, while gradual drifting on straight roads might not be, which might result from alignment issues or even a consequence of cyber-attacks on the vehicle. The group configuration is also shown in Figure \ref{fig:anomaly_composition}. 
        In this evaluation, TRAP performed better in 3 of the 4 different cases while achieving very close performance to Isolation forest in 2/5 Majority removal configuration.
    
\begin{table*}[htpb!]
\centering
\caption{F1-Scores for Different Group Removal Configurations}
\label{tab:group_removal}
\resizebox{0.8\textwidth}{!}{%
\begin{tabular}{@{}lcccc@{}}
\toprule
\textbf{Models} & \textbf{2/5 Majority removal} & \textbf{3/5 Majority removal} & \textbf{3/5 Minority removal} & \textbf{2/5 Minority removal} \\ \midrule
OCSVM  & 0.719 & 0.719 & 0.657 & 0.629 \\
IForest  & \textbf{0.889} & 0.849 & 0.801 & 0.692 \\
LOF      & 0.727 & 0.725 & 0.723 & 0.731 \\
LSTM (Trained WCT) & 0.416 & 0.736 & 0.611 & 0.617 \\
\hline
\textbf{Ours **}   & 0.869 & \textbf{0.871} & \textbf{0.820} & \textbf{0.732} \\
\bottomrule
\end{tabular}%
}
\end{table*}

\begin{table}[ht!]
\centering
\caption{Impact of noise addition on trajectory displacement}
\resizebox{0.45\textwidth}{!}{ %

    \begin{tabular}{lccccccc}
    \toprule
    \textbf{Noise Type} & \textbf{Intensity} & \textbf{MSE} & \textbf{MAE} & \textbf{Mean Displacement} & \textbf{DTW Distance} \\
    \midrule
    \multirow{3}{*}{\textbf{Precipitation}}  & Low  & 0.000433 & 0.000453 & 0.116576 & 34.97262 \\
                                    & Med  & 0.002634 & 0.002757 & 0.516898 & 155.06934 \\
                                    & High & 0.007308 & 0.007664 & 0.955337 & 286.60100 \\
    \midrule
    \multirow{3}{*}{\textbf{Distance}}       & Low  & 0.000469 & 0.000703 & 0.124870 & 37.460857 \\
                                    & Med  & 0.000470 & 0.001015 & 0.130962 & 39.288517 \\
                                    & High & 0.000473 & 0.001563 & 0.142240 & 42.672092 \\
    \midrule
    \multirow{3}{*}{\textbf{Ambient}}        & Low  & 0.022225 & 0.027014 & 2.145561 & 643.6665 \\
                                    & Med  & 0.066651 & 0.081020 & 3.772315 & 1131.6945 \\
                                    & High & 0.111125 & 0.135086 & 4.885277 & 1465.5833 \\
    \midrule
    \multirow{3}{*}{\textbf{Spurious}}       & Low  & 0.000167 & 0.002498 & 0.211202 & 63.360645 \\
                                    & Med  & 0.001998 & 0.014988 & 0.740258 & 222.077550 \\
                                    & High & 0.007499 & 0.037496 & 1.437241 & 431.172200 \\
    \midrule
    \multirow{3}{*}{\textbf{Composite1}}      & Low  & 0.02321 & 0.030438 & 2.244333 & 673.3000 \\
                                    & Med  & 0.07130 & 0.097181 & 4.041133 & 1212.3397 \\
                                    & High & 0.12451 & 0.171568 & 5.434584 & 1630.3752 \\
    \midrule
    \multirow{3}{*}{\textbf{Gaussian}}       & Low  & 0.000524 & 0.004846 & 0.216445 & 64.93355 \\
                                    & Med  & 0.000688 & 0.009220 & 0.323226 & 96.96787 \\
                                    & High & 0.001343 & 0.017945 & 0.549140 & 164.74188 \\
    \midrule
    \multirow{3}{*}{\textbf{Dropout}}       & Low  & 0.042195 & 0.043958 & 2.415912 & 724.7734 \\
                                    & Med  & 0.126722 & 0.132006 & 4.202101 & 1260.6304 \\
                                    & High & 0.211215 & 0.220021 & 5.428938 & 1628.6816 \\
    \midrule
    \multirow{3}{*}{\textbf{Composite2}}       & Low  & 0.042885 & 0.050559 & 2.551467 & 765.4401 \\
                                    & Med  & 0.129203 & 0.150395 & 4.605394 & 1381.6184 \\
                                    & High & 0.219454 & 0.255462 & 6.214792 & 1864.4377 \\
    \midrule
    \end{tabular}
    }
\end{table}

\section{Conclusion}

In this paper, we introduced a novel reward-guided pretraining approach for detecting anomalous driving behaviors in autonomous vehicles. Our method surpasses conventional and recent IRL-based anomaly detection techniques, demonstrating strong generalization even when trained on expert trajectories with biased anomaly composition. By leveraging worst-case terminus training, we jointly optimize extrapolated rewards with variable-horizon trajectory sampling, implicitly maximizing the time of consequence while reshaping the reward structure within the embedder. This eliminates the need for extensively labeled anomaly datasets while achieving superior performance. Comprehensive evaluations across diverse noise configurations and anomaly-grouped expert training validate the real-world applicability of our approach. Additionally, our method provides interpretable insights into the reward structures guiding driving behavior. Future work can extend this framework to more complex driving scenarios.

\section*{Acknowledgments}
This work was supported by Clemson University’s Virtual Prototyping of Autonomy Enabled Ground Systems (VIPR-GS), under Cooperative Agreement W56HZV-21-2-0001 with the US Army DEVCOM Ground Vehicle Systems Center (GVSC).

DISTRIBUTION STATEMENT A. Approved for public release; distribution is unlimited. OPSEC\# 9830.

\bibliographystyle{IEEEtran} 
\bibliography{references} 
\end{document}